\documentclass{article}
\usepackage{spconf,amsmath,graphicx}
\usepackage{amsmath}
\usepackage{amsfonts}

\usepackage{algorithm}
\usepackage{algorithmic}
\usepackage[compatibility=false]{caption}
\usepackage{multirow}
\usepackage{subcaption}
\usepackage{amsmath}
\usepackage[noadjust]{cite}
\usepackage{pifont}

\newcommand{\dcn}{{\textsc{DualCoreNet }}}

\newcommand{\dcnn}{{\textsc{DualCoreNet}}}

\title{A deep dual-path network for Improved Mammogram Image Processing}
\name{Heyi Li, Dongdong Chen, William H. Nailon, Mike E. Davies and Dave Laurenson}
\address{School of Engineering, The University of Edinburgh, Edinburgh, UK.
}

\begin{document}
\maketitle
\begin{abstract}

We present, for the first time, a novel deep neural network architecture called \dcn with a dual-path connection between the input image and output class label for mammogram image processing. This architecture is built upon U-Net, which non-linearly maps the input data into a deep latent space.  One path of the \dcnn, the locality preserving learner, is devoted to hierarchically extracting and exploiting  intrinsic features of the input, while the other path, called the conditional graph learner, focuses on modeling the input-mask correlations. The learned mask is further used to improve classification results, and the two learning paths complement each other. By integrating the two learners our new architecture provides a simple but effective way to jointly learn the segmentation and predict the class label. Benefiting from the powerful expressive capacity of deep neural networks a more discriminative representation can be learned, in which both the semantics and structure are well preserved. Experimental results show that \dcn achieves the best mammography segmentation and classification simultaneously, outperforming recent state-of-the-art models. 
\end{abstract}

\section{Introduction}
\label{sec:intro}
According to the International Agency for Research on Cancer \cite{IARC2008}, breast cancer is the most frequently diagnosed cancer and the second most fatal disease among women around the world. Breast masses are the most challenging breast abnormality to diagnose and the most feared, since about 90\% of breast masses are cancerous. Although there are no effective methods for prevention, early intervention is critical for improving associated survival rates. Screening mammography is thus employed for invasive malignant tumours (measuring $<2$ cm) when they are too small to be palpable or cause symptoms. However, the manual inspection of screening mammograms by radiologists is tedious, subjective, and prone to errors, which may cause high false positive rates and over diagnosis \cite{varela2006use,arevalo2016representation,li2018improved}. For these reasons, automatic and robust diagnosis tools for screening mammography are in high demand. Traditional mammographic computer-aided diagnosis (CAD) systems rely heavily on elaborate hand-crafted features \cite{varela2006use}. Recently, leveraging the insights from the successes of deep neural networks (deep learning) in computer vision tasks \cite{zhu2017deep,chen2018learning,carneiro2017automated,shams2018deep,golbabaee2018deep}, deep learning based algorithms have been applied to mammograms and have achieved state-of-the-art results for mass detection, segmentation and classification. Mass detection aims to find the regions of interest (ROIs) where abnormalities may be located, and mass segmentation provides detailed morphological features with precise outlines within ROIs. Mass classification categorizes mammogram patches or full mammograms into benign or malignant, assisting radiologists to interpret screenings locally (region-based mass classification) or globally (full mammogram classification). There are several recent examples which deep learning has been applied to full mammogram classification \cite{zhu2017deep,carneiro2017automated,shams2018deep}. However, these works ignore the importance of local ROIs, since radiologists are more likely to identify local regions rather than labelling the full mammograms in clinical practice. In terms of region-based mass classification, Convolutional Neural Networks (CNNs) have been applied as a classifier showing that deep learning models outperform traditional methods significantly \cite{arevalo2016representation, kooi2017discriminating}. The work in \cite{dhungel2016automated} utilized the CNN as a regressor to extract hand-crafted features from both mammogram ROIs and annotation masks of radiologists, which were then followed by a Random Forest classifier model. Although \cite{dhungel2016automated} results in better performance than direct CNN based algorithms, hand-crafted features are still required in the final design. 

\begin{figure}[t]
	\begin{center}
	    	\includegraphics[width=0.96\linewidth]{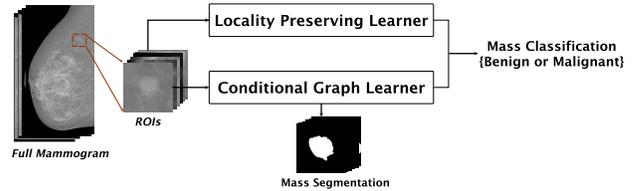}
	    	\vskip -0.1in
	\caption{The illustration of dual-path learning for mammography analysis.}
	\label{flowchart}
	\end{center}
	\vskip -0.3in
\end{figure}

\begin{figure*}
	\centering
	\includegraphics[width=0.95\textwidth]{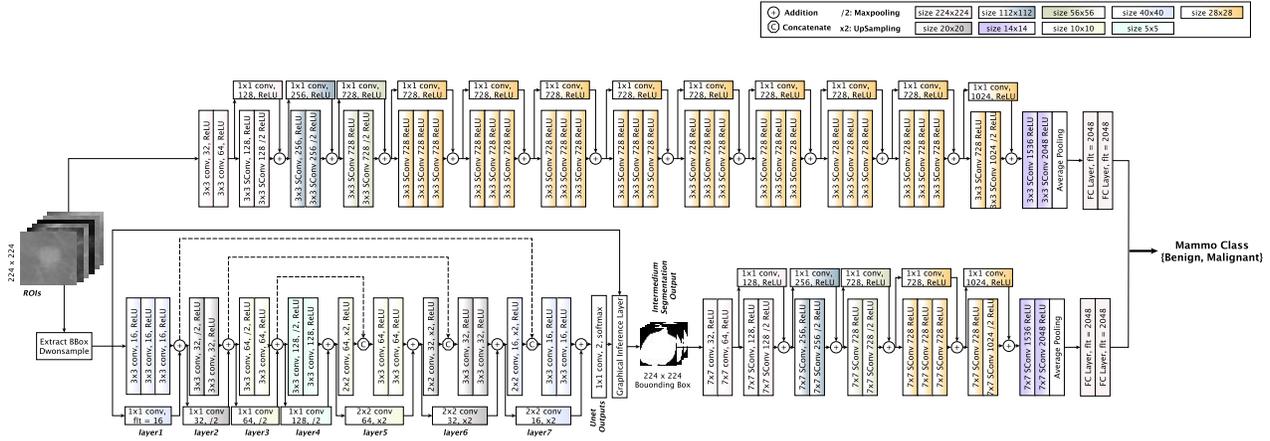}
	\caption{The proposed \textbf{Dual}-path \textbf{Co}nditional \textbf{Re}sidual \textbf{Net}work (\dcn).}
	\label{segcls}
	\vskip -0.1in
\end{figure*}

More recently, \cite{ronneberger2015u} proposed a powerful CNN based image segmentation model referred as U-Net, which interlaces multi-resolution information by adding skips between encoding and decoding layers of the same spatial size and has been shown to perform well on medical images. However, when applied to mammography, the U-Net is limited by low-signal-to-noise data, which causes labelling inconsistency and incompleteness. To address this, a novel architecture is designed in this paper. Based on the accurate pixel-level labelling algorithm presented in \cite{li2018improved} and the related breast mass classifiers \cite{zhu2017deep,carneiro2017automated,shams2018deep,arevalo2016representation, kooi2017discriminating, dhungel2016automated, varela2006use}, a \textbf{Dual}-path \textbf{Co}nditional \textbf{Re}sidual \textbf{Net}work (\dcnn) for mammography analysis is proposed as shown in Figure \ref{flowchart}. Firstly, a mass and its context texture learner called the \textit{Locality Preserving Learner (LPL)} is built with stacks of convolutional blocks, achieving a mapping from ROIs to class labels. Secondly, a graph inference layer, called the \textit{Conditional Graph Learner (CGL)} is employed to learn the input-mask correlation, and the extracted segmentation features will be further used to improve the final mass classification performance. By integrating these two learning paths, \dcn achieves the best mammography segmentation and classification simultaneously, outperforms recent state-of-the-art models. The main contributions of this paper are the following: (i) To our knowledge, \dcn is the first dual-path CNN-based mammogram analysis model that takes advantage of segmented mask for mass classification; (ii) Our model achieves state-of-the-art results on both mass segmentation and mass classification tasks on publicly available mammography datasets. 

\section{Methodology}
\label{sec:met}
The \dcn takes mammogram ROIs as input and outputs mass segmentation masks and ultimately classifies ROIs simultaneously. The pipeline of \dcn is shown in Fig \ref{segcls}. We first define the notation used throughout the paper and then introduce the details of the two proposed learning paths and the whole \dcnn.

\subsection{Notations}
Given a mammogram lattice $\mathrm{\Omega}$, let $\boldsymbol{x}: \mathrm{\Omega}\rightarrow\mathbb{R}$ as one of the $N$ training mammogram ROIs, $\boldsymbol{y}_{seg}: \mathrm{\Omega}\rightarrow\{0: \text{Normal pixels}, 1:\text{Mass pixels}\}$ as the corresponding pixel-level annotations from radiologists  and $\boldsymbol{y}_{cls} \rightarrow\{0:\text{Benign}, 1:\text{Malignant}\}$ for the ROI class labels, the training set is represented by $\mathcal{D}=\{(\boldsymbol{x}^{(n)}, \boldsymbol{y}_{seg}^{(n)}, \boldsymbol{y}_{cls}^{(n)} )\}_{n=1}^N$.

\subsection{Locality Preserving Learner}
Mass and its context tissue textures are exploited as the major classification features in traditional mammographic CADs \cite{varela2006use}. The LPL learns the hierarchical and local intrinsic texture features by non-linear mapping locality information's linear combination with stacks of depth-wise separable convolution, maxpooling and ReLU activation. 

In order to prevent the model from exploding and vanishing gradients, the residual learning \cite{he2016deep} is employed within the whole \dcnn, which maps the convolutional layer's output $\mathcal{H}(\boldsymbol{x})$ with residuals $\mathcal{F}(\boldsymbol{x})$ and the linear dimension matching kernel $W$ as:
\begin{equation}
\mathcal{F}(\boldsymbol{x}):=\mathcal{H}(\boldsymbol{x})-W*\boldsymbol{x}
\end{equation}  

Depth-wise separable convolutions, referred as "SConv" in Fig \ref{segcls}, are widely applied to efficiently map the network by factoring filters into a series of operations, such as \cite{chollet2017xception}. Specifically, depth-wise separable convolution comprises of a spatial convolution that operates independently for each feature map and a pointwise convolution that computes a linear projection for the spatial convolution's output. Inspired by this, residual learning paths with the depth-wise separable convolutions are built in the \dcnn to independently learn cross-channel and spatial correlations of hierarchical but local intrinsic features from both mammogram ROIs and segmentation masks. The loss associated to the LPL layer is defined with categorical cross-entropy as:
\begin{equation}\label{eq:lpl}
    \ell_{lpl} = -\sum_{n=1}^N\log P\big({y_{cls}}^{(n)} \mid \boldsymbol{x}^{(n)};\boldsymbol{\theta}_{lpl}\big)
\end{equation}
where $y_{cls}$ is the class indicator, $\boldsymbol{\theta}_{lpl}$ is the parameter set of the LPL path. 

\subsection{Conditional Graph Learner}
In practice, radiologists usually analysis mass shape and boundary information to improve cancer inspections, the more irregular the shape, the more likely to be cancerous \cite{dhungel2015deep}. This observation suggests that extracting segmentation-friendly features should improve mass classification performance. To do that, the CGL is proposed accordingly. Specifically, the CGL first employs the revised U-Net, mapping ROIs to geometry and spatial features, then a graphical inference layer to preserve pixel consistency with conditional restrictions and finally the separable convolution stacks to efficiently learn hierarchical features and class labels mapping. 


Firstly, the U-Net makes use of standard convolutional layers and employs a combination of multi-resolution filters. Incorporated with residual learning in the CGL, the $m$th layer output with input $\boldsymbol{x}^{(n)}$ at pixel $(i,j)$ is formulated as:
\begin{equation}
\begin{split}
{y}^{(n,m)}_{i,j} &= \sigma(h_{ks}(\{\boldsymbol{x}_{s_i+\delta_i, s_j+\delta_j}\}_{0\leq\delta_i,\delta_j\leq k})) \\
&+ \sigma(h'_{k's'}(\{\boldsymbol{x}_{s'_i+\delta_i, s'_j+\delta_j}\}_{0\leq\delta_i,\delta_j\leq k'}))
\end{split}
\label{layer_func}
\end{equation}
where $h_{ks}$ and $h'_{k's'}$ denote the convolution functions of residuals and dimension mapping respectively, $k$ and $k'$ represent the corresponding kernel sizes of convolutional filters, $s$ and $s'$ for kernel strides or maxpooling factors in the downsampling layers, and $\sigma$ for activation functions.

The graphical inference layer applies conditional random fields (CRFs) as an additional CNN layer \cite{li2018improved,zheng2015conditional}. The loss function for the segmentation is defined with weighted categorical cross-entropy of the residual learning U-Net and graphical inference layer as follows
\begin{equation}
\begin{split}
\ell_{seg} &=(1-\lambda)\Big(-\sum_{i,j}\log P\big({y}_{i,j}^{(n)}\mid\boldsymbol{x}^{(n)};\boldsymbol{\theta}_{seg}\big)\Big)\\
+&\lambda \cdot \Big( A(\boldsymbol{x}^{(n)}) 
- \exp\big(\sum_{i,j\in V} P\big({y}_{i,j}^{(n)}\mid\boldsymbol{x}^{(n)};\boldsymbol{\theta}_{seg}\big) \\
+&\sum_{p,q\in{E}}\phi\big(y_{p}^{(n)},y_q^{(n)} \mid\boldsymbol{x}^{(n)}\big)\Big)
\label{loss_unet}
\end{split}
\end{equation}
where $P({y^{(n)}_{i,j}})$ is the residual U-Net output probability distribution at position $(i,j)$ given the parameters $\boldsymbol{\theta}_{seg}$, $A$ is the partition function of the CRF, $\phi$ is the pair-wise potential function which is defined with the label compatibility, Gaussian kernels and corresponding weights for pixel $p$ and $q$ belonged to the CRF graph edges $E$, and $\lambda$\footnote{Optimally chosen as 0.67 with grid search in our experiments.} is the trade-off factor. 

The loss of the CGL path is again defined by the classification categorical cross-entropy as:
\begin{equation}\label{eq:cgl}
    \ell_{cgl} = -\sum_{n=1}^N\log P\big({y_{cls}}^{(n)}\mid{P(\boldsymbol{y}_{seg}^{(n)})};\boldsymbol{\theta}_{cgl}\big) 
\end{equation}
where $ P(\boldsymbol{y}_{seg}^{(n)})$ is the resulted soft mass mask and $\boldsymbol{\theta}_{cgt}$ is the parameter set of the CGL path. 

Finally, by integrating CGL and LPL into a dual-path network model, the overall categorical cross-entropy based loss of the \dcn is defined as:
\begin{equation}\label{eq:dcn}
\ell = -\sum_{n=1}^N\log P\big({y_{cls}}^{(n)} \mid \boldsymbol{x}^{(n)};\boldsymbol{\theta}_{cgl}, \boldsymbol{\theta}_{lpl} \big)
\end{equation}

\section{Experiments}
\label{sec:exp}
\subsection{Datasets and ROIs selection}
\label{ssec:data}
The \dcn is evaluated on two public datasets: INbreast \cite{moreira2012inbreast} and CBIS-DDSM \cite{lee2016curated}. 
INBreast is a full-field digital mammographic (FFDM) dataset containing 116 pixel-level annotated masses in 107 mammograms, with pixel size $70 \mu m$ and contrast resolution 14 bits. The CBIS-DDSM is a modernized subset of Digital Database for Screening Mammography (DDSM) \cite{heath2000digital}, that includes 1594 mass contained digitized film screening mammograms.

Two manual ROIs selection procedures are adopted for the training stage in this work: One is to locate and extract rectangular mass contained bounding boxes, and they are utilized by the CGL to explore the boundary and shape features; the other ROI set is selected with proportional padding that mass-centred ROIs include regions $1.6$ times the size of the mass bounding box, and they are utilized by the LPL for mass and its context tissues texture feature learning. The selected ROIs are then augmented with horizontal and vertical flips. In terms of data division, the INbreast data set is divided by patients into a training set and a test set as 80\%: 20\%. As for the CBIS-DDSM, we adopted the pre-divided 1253 training and 354 test ROIs within the data. 

\subsection{Implementation Details}
As shown in Fig \ref{segcls}, the LPL path is designed with stacks of separable convolutional layers with kernel size $3\times3$ for fine texture features learning. The first three blocks are equipped with increasing feature maps (128, 256, 728) and decreasing spatial size ($224\times224$, $112\times112$, $56\times56$), while the consecutive eight blocks are of the same feature map (728) and the same spatial shape ($28\times28$). 

The CGL path takes downsampled bounding box ROIs (with spatial size $40 \times 40$) as the input and followed by four consecutive maxpooling and convolution blocks with feature map numbers 16, 32, 64 and 128 and corresponding spatial size $40\times40$, $20\times20$, $10\times10$ and $5\times5$. The bottom latent features are then upsampled reversely with convolution operations and skips that connect previous layer channels. The multi-resolution mapping is then followed by the CRF graphical inference layer, resulting in a sigmoid activated upsampled segmentation mask (size $224\times224$). A relative large kernel $7\times7$ is chosen for the appended CNN blocks, aiming to better learn the shape and boundary features. 

After global averaging on both paths and two fully connected layers with 2048 neurons separately, the concatenated features of dual paths are classified into either one of the two mass categories (Benign or Malignant). Except for the separable convolutional blocks in the LPL path, which are initialized with ImageNet \cite{russakovsky2015imagenet} pretrained weights for an accelerate convergence rate and network generalization, all other network layers were randomly set. To avoid over-fitting, dropout layers with $50\%$ dropout rate were used. The \dcn is optimized by the Stochastic Gradient Descent algorithm with the Adam update rule. To fully train the \dcnn, two component paths were first trained with their corresponding loss function (\ref{eq:lpl}) and (\ref{eq:cgl})  separately, and then further trained jointly with the loss (\ref{eq:dcn}). 
\begin{table}[t]
	\renewcommand{\arraystretch}{1}
	\caption{Mass segmentation performance (within Bounding Box ROIs) of the \dcn and several state-of-the-art methods on test datasets.}
	\label{seg_results}
	\centering
	\begin{tabular}{c|c|c}
		\hline
		\bfseries Methodology & \mdseries Dataset  & \mdseries DI, \% \\
		\hline 
		Dhungel \textit{et. al.} \cite{dhungel2015deep}& INbreast  & $90$ \\
		Zhu \textit{et. al} \cite{zhu2018adversarial}& INbreast  & $89.36$ \\
		\dcn & INbreast & $\mathbf{93.66}$ \\
		\hline
		Dhungel \textit{et. al.} \cite{dhungel2015deep}& CBIS-DDSM  & $90$ \\
		Zhu \textit{et. al} \cite{zhu2018adversarial}& CBIS-DDSM  & $90.62$ \\
		\dcn & CBIS-DDSM & $\mathbf{91.43}$ \\
		\hline
	\end{tabular}
\end{table}   

\begin{figure}[t]
	\begin{center}
	    	\includegraphics[width=0.87\linewidth]{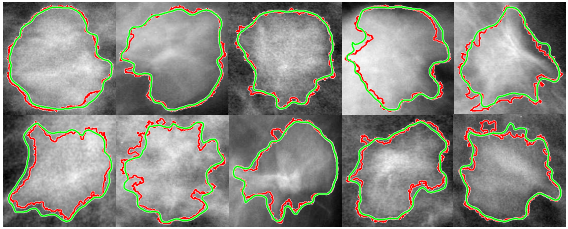}
	\caption{ Visualized comparison between radiologists' annotation (red lines) and \dcn segmentation results (green lines).}
	\label{contours}
	\end{center}
	\vskip -0.13in
\end{figure}

\subsection{Results}
In terms of mass segmentation, evaluations of Dice Coefficients are compared with related works in Table \ref{seg_results}. \dcn achieves the best performance with 93.66\% and 91.43\% on INbreast and CBIS-DDSM, respectively. Visualized overlapping contours of the radiologist's annotation (red lines) and \dcnn's segmentation results (green lines) are shown in Figure \ref{contours}. The \dcn contributes similar to the radiologists and smooth contours even for relative hard cases. 

\begin{figure}[t]
	\normalsize
	\begin{center}
    	\includegraphics[width=0.985\linewidth]{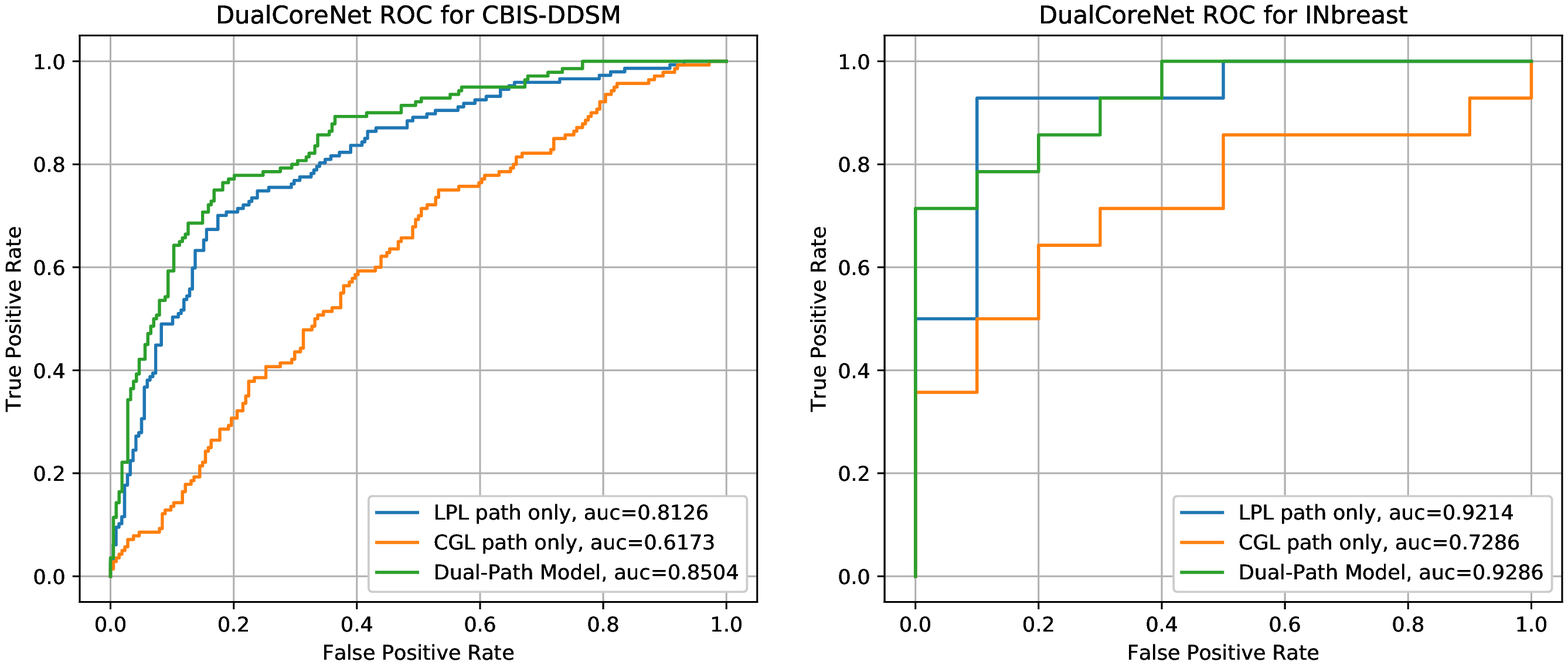}
    	\vskip -0.16in
		\caption{Mass classification ROC curves of the \dcn experimented on INbreast (left, 85 training and 22 test mammograms) and CBIS-DDSM (right, 1253 training and 354 test mammograms).}
	\label{rocs}
	\end{center}
	\vskip -0.25in
\end{figure}

As for the mass classification performance, the ROC curve and AUC of LPL, CGL and joined paths' best model are shown in Figure \ref{rocs}. For INbreast, the dual path model achieved the best performance 0.93 AUC, 0.01 higher than LPL path only. With a guaranteed classification performance, the dual path model further contributes precise mass segmentation. For the CBIS-DDSM, the dual-path model outperforms either CGL or LPL, achieving 0.85 AUC. Moreover, as shown in Table \ref{table_results}, the \dcn outperforms other state-of-art algorithms. The experimental results show that the representation learned by \dcn is more robust for both of breast mass segmentation and classification in mammography, which again demonstrate our motivation of dual-path learning.

\begin{table}[t]
	\renewcommand{\arraystretch}{1}
	\caption{Region based mass classification performance (AUC) of the proposed dual paths model and relative state-of-the art methods on test sets.}
	\label{table_results}
	\centering
	\begin{tabular}{c|c|c}
		\hline
		\bfseries Methodology & \mdseries  CBIS-DDSM & \mdseries INbreast  \\ 
		\hline
		 Dhungel \textit{et. al.} \cite{dhungel2016automated}& - &  $0.91$  \\
		 Varela  \textit{et. al.} \cite{varela2006use}   & $0.83$ & -  \\
		 \hline 
		 \dcn (CGL only) & $0.62$ & $0.73$ \\
		 \dcn (LPL only) & $0.81$ & $0.92$\\
		 \dcn (Dual-path) & $\mathbf{0.85}$ & $\mathbf{0.93}$\\
		\hline 
	\end{tabular}
\end{table}

\section{Conclusions}

In this paper, we proposed a novel \dcn for improved mammogram image analysis. By integrating the conditional graph learner path and the locality preserving learner path,  our \dcn works in a simple but effective way to jointly learn segmentation and classification. Thanks to the departure from hand-crafted features in classical CADs, our model is more flexible to learn mammographical-friendly representations. Furthermore, \dcn performs better on higher quality dataset (the INbreast dataset). Extensive experiments show that our method outperforms state-of-the-art on both breast mass segmentation and classification tasks in mammography. 

\newpage
\bibliographystyle{IEEEtran}
\bibliography{strings,refs}

\end{document}